\documentclass{article}

    \usepackage[preprint]{neurips_2024}

\usepackage[utf8]{inputenc} 
\usepackage[T1]{fontenc}    
\usepackage{hyperref}       
\usepackage{url}            
\usepackage{booktabs}       
\usepackage{amsfonts}       
\usepackage{nicefrac}       
\usepackage{microtype}      
\usepackage[dvipsnames]{xcolor}         

\usepackage{amsmath, scalerel}
\usepackage{subcaption}
\usepackage{graphicx}
\usepackage{graphbox}
\usepackage[inkscapelatex=false]{svg}
\usepackage{wrapfig}

\newtheorem{theorem}{Proposition}
\DeclareMathOperator*{\concat}{\scalerel*{\Vert}{\sum}}
\title{Graph Transformers without Positional Encodings}

\author{%
  Ayush~Garg\thanks{Independent research} \\
  \texttt{ayush.garg@alumni.ethz.ch} \\
}

\begin{document}

\maketitle

\begin{abstract}
  Recently, Transformers for graph representation learning have become increasingly popular, achieving state-of-the-art performance on a wide-variety of graph datasets, either alone or
  in combination with message-passing graph neural networks (MP-GNNs). Infusing graph 
  inductive-biases in the innately structure-agnostic transformer architecture in the 
  form of structural or positional encodings (PEs) is key to achieving these impressive results.
  However, designing such encodings is tricky and disparate attempts have been made 
  to engineer such encodings including Laplacian eigenvectors, relative random-walk probabilities (RRWP), spatial encodings, centrality encodings, edge encodings etc. In this work, we argue that such encodings may not be required at all, provided the attention mechanism itself incorporates information about the graph structure. We introduce \textsc{Eigenformer}, a Graph Transformer employing a novel \emph{spectrum-aware} attention mechanism cognizant of the Laplacian 
  spectrum of the graph, and empirically show that it achieves performance competetive with SOTA Graph Transformers on a number of standard GNN benchmarks. Additionally, we theoretically prove that \textsc{Eigenformer} can express various graph structural connectivity matrices, which is particularly essential when learning over smaller graphs.
\end{abstract}

\section{Introduction}
Learning useful representations from graph data is important in a variety of domains due to the ubiquitous presence of graphs around us, ranging from molecules to social and transportation networks. It may well be argued that "graphs are the main modality of data we receive from nature" \citep{EverythingConnected}. In general, machine learning methods designed to learn patterns from graph data fall into two paradigms: message-passing graph neural networks (MP-GNNs), where vector messages are exchanged between neighboring nodes and updated using neural networks \citep{Bruna, Quantum}, and more recently Transformers \citep{Vaswani} --originally designed for Natural Language Processing (NLP) tasks-- for graph data, which feature long-range connections between nodes for faster information exchange albeit with a loss of graph inductive biases \citep{DwivediBresson,SAN,StructureAware,Ying,GraphGPS,Hussain,zhang2023rethinking}.

In many ways, MP-GNNs and Graph Transformers have complementary strengths and weaknesses. While MP-GNNs have strong graph inductive biases due to sharing of messages within local neighborhoods, they suffer from issues such as over-squashing \citep{alon2021on,topping2022understanding}, over-smoothing \citep{deeper,Oono2020Graph} and expressivity concerns (upper bounded by the 1-Weisfeiler-Lehman isomorphism test) \citep{xu2018how,Loukas2020What,WL}. On the other hand, Graph Transformers do not suffer from these issues, at the expense of losing graph connectivity information (forgoing locality in favor of all-to-all connectivity between nodes), possibility of over-fitting and quadratic computational and memory complexity. Due to their complementary benefits, researchers have (successfully) tried to merge the two paradigms together into one architecture with GraphGPS in \citet{GraphGPS}. 

A common weakness in both methods, however, is the lack of node positional information, which has been shown to be crucial both to improve expressivity in MP-GNNs (as compared to the 1-WL test in \citet{xu2018how}), and to allow Transformers to be used with datasets with smaller graphs. Nodes in a graph lack a canonical ordering, and superficially imposing an ordering leads to learning difficulties due to the exponential growth in such possible orderings. To tackle this, positional encodings (PEs) based on Laplacian eigenvectors were proposed in \citet{DwivediBresson}. Due to sign ambiguities of eigenvectors and eigenvalue multiplicities, sign and basis invariant/equivariant methods such as SignNet and BasisNet in \citet{SignNet} were proposed. In other works \citep{RWPE,GRIT}, \(k\)-dimensional (learnable) random-walk probabilities from a node to itself are used instead of Laplacian eigenvectors to avoid the aforementioned invariance issues.

\textbf{Main contributions:} In light of the above background, we ask the question whether such encodings are indeed necessary at all. If there is no canonical node-ordering inherent to a graph, perhaps no such ordering is required to effectively learn patterns from graph data. Thus, in this work, we seek to circumvent the issues related to designing node/edge PEs by modifying the way node feature information is exchanged in a Graph Transformer. Specifically:

\begin{itemize}
    \item Instead of adding positional information to node features and letting the attention mechanism figure out the strength of connection between a pair of nodes, we factorize the attention matrix in terms of fixed node-pair "potentials" and learned frequency "importances". We posit such a factorization encodes important graph inductive biases from the frequency domain into the attention mechanism while being flexible enough to yield impressive performance across domains and objectives.
    \item We utilize valuable information from both the eigenvectors and the eigenvalues as advocated in \citet{SAN}, forgoing the assumption that higher frequencies are less important in determining node-pair interactions.
    \item The novel attention mechanism with built-in inductive biases allows us to train wider/deeper models for a given parameter budget due to simpler Transformer layers. Numerical experiments demonstrate the strength of our method, achieving near-SOTA performances on many common GNN benchmarks.
\end{itemize}

\section{Theoretical Motivations}
We first present the motivation behind the architecture of \textsc{Eigenformer} by briefly reviewing the semantics of the graph Laplacian, its spectrum and spectral convolution.

\subsection{Graph Laplacian and its spectrum}

For an undirected graph \(\mathcal{G} = (\mathcal{V}, \mathcal{E})\), the unnormalized graph Laplacian \(L\) is defined as
\begin{equation}\label{L}
    L = D - A
\end{equation}
where \(A\) is the adjacency matrix and \(D\) is the degree matrix. 
By definition \ref{L}, the following is true for a unit-norm eigenvector \(u_i\) and its corresponding eigenvalue \(\lambda_i\):
\begin{equation}\label{lambda}
    \lambda_i = u_i^TLu_i = \frac{1}{2}\sum_{u,v \in \mathcal{V}}A[u,v](u_i[u]-u_i[v])^2 = \sum_{(u,v)\in\mathcal{E}}(u_i[u]-u_i[v])^2
\end{equation}

If the eigenvectors are interpreted as "signals" over the graph, taking real scalar values at nodes, then the eigenvalues act as estimates of the "smoothness" of such signals, measuring how much the signal values differ between endpoints of edges. The smallest eigenvalue is always 0, assigning the same scalar value \(u_i[\cdot]\) to all nodes in the same connected component of the graph. 

\begin{figure}[t!]
\centering
\includegraphics[]{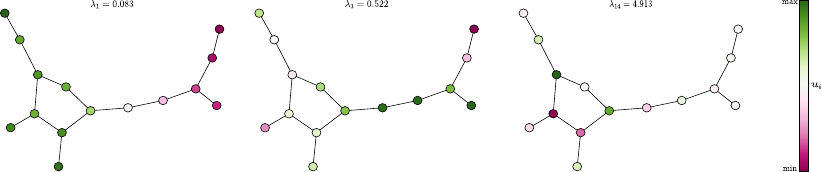}
\caption{Example molecule from PCQM4Mv2 dataset: Substructures are revealed by eigenvectors}
\label{fig:substructures}
\end{figure}

Thus, eigenvectors of the Laplacian encode important information about node similarities (nodes resonant under a frequency tend to have similar values w.r.t its eigenvector). As such, the spectrum --in its interpretation as frequencies of resonance of the graph-- discriminates between different graph structures and substructures. Figure \ref{fig:substructures} shows substructures revealed by eigenvectors for an example molecular graph from the PCQM4Mv2 dataset \citep{ogb_lsc} containing 15 nodes. Color-codes indicate the value \(u_i[\cdot]\) for different frequencies \(\lambda_i \) where \(i \in \{0,..., 15\}\). Lower frequencies highlight larger substructures while higher frequencies emphasize more local interactions.

\subsection{Spectral convolution} \label{specconv}

Since the (sinusoidal) eigenfunctions \(f(t) = e^{2\pi ist}\), of the general Laplace operator \(\Delta\), correspond to the modes of the frequency domain of the Fourier transform, the eigenvectors of the graph Laplacian \(L\) can be equated to sine functions over graphs. This connection allows us to generalize the Fourier transform to arbitrary graphs. 

For a signal \(f \in \mathbb{R}^{\vert \mathcal{V} \vert}\) on an undirected graph \(\mathcal{G} = (\mathcal{V},\mathcal{E})\) with Laplacian \(L=U \Lambda U^T\), where \(U\) is the orthonormal matrix of eigenvectors and \(\Lambda\) is the diagonal matrix of eigenvalues, we can compute its Fourier transform as \(s = U^Tf\), and its inverse as \(f = Us\). Thus, its graph convolution \(\star_\mathcal{G}\), with a filter \(h\) can be computed as:

\begin{equation}
    f \star_\mathcal{G} h = U (U^Tf \odot U^Th) = (U \text{diag}(w_h) U^T)f
\end{equation}

where the last equality is true when the filter \(h\) can be learnt in a non-parametric way by representing and optimizing directly in the spectral domain: \(w_h = U^Th\). Such a filter, however, may not satisfy sound graph convolution properties like locality and translational equivariance, and it is useful to parameterize \(w_h\) in terms of eigenvalues \(\Lambda\) of the Laplacian, for example, as a polynomial \(p(\Lambda)\), which ensures commutativity with the Laplacian, restoring the required properties:

\begin{equation}
    f \star_\mathcal{G} h = (U p(\Lambda) U^T)f = p(L)f
\end{equation}

Thus, a spectral convolution on the graph can be defined in terms of a polynomial of the Laplacian \(L\). The degree \(k\), of the polynomial determines the \(k\)-hop neighborhood over which the convolution is performed.

\section{Model Architecture}
In this section, we introduce our novel \emph{spectrum-aware} attention (SAA) mechanism that builds on the spectral convolution theory discussed in \ref{specconv} and demonstrate its use in the proposed \textsc{Eigenformer} architecture. 

\textbf{Note:} We use the normalized variant of the graph Laplacian, \(L_{norm} = D^{-\frac{1}{2}}LD^{-\frac{1}{2}}\) that admits similar properties and interpretations of the spectrum as that of \(L\), but has better numerical stability due to a bounded spectrum: all eigenvalues of \(L_{norm}\) lie between \(0\) and \(2\). Moreover, the normalized Laplacian counters the effect of "heavy" nodes (nodes with high degrees) in influencing information propagation through the graph. Also, for directed graphs, we first make the edges undirected before calculating \(L_{norm}\) so that its spectrum remains real.

\subsection{Attention using the Laplacian spectrum}\label{AttentionLaplacian}

For a graph \(\mathcal{G}=(\mathcal{V}, \mathcal{E})\) 
we define the attention weight between two nodes \(i\) and \(j\) in \(\mathcal{G}\), through the following equations:

\begin{equation}\label{sigma}
\sigma_k[i,j] =u_k[i] \cdot u_k[j]
\end{equation}

\begin{equation}\label{alpha}
\alpha[i,j] = softmax_{j \in \mathcal{V}}\Bigl[\phi_1\Bigl(\sum_{k=1}^{\vert\mathcal{V}\vert}\sigma_k[i,j]\phi_2(\lambda_k)\Bigl)\Bigl]
\end{equation}

where \(u_k \in \mathbb{R}^{\vert\mathcal{V}\vert}\) 
is the eigenvector corresponding to the eigenvalue \(\lambda_k \in \mathbb{R}\) of \(L_{norm}\), \(\sigma_k \in \mathbb{R}\) can loosely be interpreted as the "potential" between the nodes under the frequency \(\lambda_k\) (see Green's function from electrostatic theory, \citet{Greens}), \(\phi_1: \mathbb{R} \rightarrow \mathbb{R}\) and \(\phi_2: \mathbb{R} \rightarrow \mathbb{R}\) are arbitrary real functions and \(\alpha \in \mathbb{R}\) is the final attention weight. The connection to spectral convolution can be inferred by noting that \(\alpha = softmax[\phi_1(U\phi_2(\Lambda)U^T)]\) where \(U = [u_1, ..., u_{\vert\mathcal{V}\vert}]\).

Propositions \ref{proposition1} and \ref{proposition2} below justify the choice of \(\sigma\) and \(\alpha\) rigorously.



\begin{theorem} \label{proposition1}
    For any \(n \in \mathbb{N}\), consider the adjacency matrix \(A\) drawn from the set of adjacency matrices of n-node undirected graphs, \(\mathbb{G}_n \subset \{0,1\}^{n \times n}\). Further, let \(L_{norm} = I - D^{-\frac{1}{2}}AD^{-\frac{1}{2}} = I - A_{norm}\) be the normalized graph Laplacian of the graph \(\mathcal{G}=(\mathcal{V},\mathcal{E})\) with \(\vert\mathcal{V}\vert=n\), adjacency matrix \(A\), eigenvalues \(\lambda_k\) and eigenvectors \(u_k\), for \(k \in \{1,...,n\}\). Then, we have the following approximations up to an arbitrary small error, \(\epsilon\):
    \begin{enumerate}
        \item \(\phi_1\big(\sum_{k=1}^{\vert\mathcal{V}\vert}\sigma_k[i,j]\cdot\phi_2(\lambda_k)\big) \approx \sum_{k=0}^{m}\theta_kA_{norm}^k[i,j]\) for \(m \in \mathbb{Z}\) 
        \item \(\phi_1\big(\sum_{k=1}^{\vert\mathcal{V}\vert}\sigma_k[i,j]\cdot\phi_2(\lambda_k)\big) \approx f(SPD[i,j])\)
    \end{enumerate}
    for suitable functions \(\phi_1\) and \(\phi_2\), where \(SPD[i,j]\) is the shortest path distance between nodes \(i\) and \(j\) and \(f\) is any continuous function.
\end{theorem}

\emph{Proof:} The proof of (\textit{1.}) utilizes the diagonalizability of the matrices \(A_{norm}\) and \(L_{norm}\)
, allowing us to derive the analytic form of \(\phi_1\) and \(\phi_2\). For (\textit{2.}), we show that \(\phi_2\) can be chosen to yield a valid (possibly many-to-one) function that maps \(\sum_{k=1}^{\vert\mathcal{V}\vert}\sigma_k[i,j]\cdot\phi_2(\lambda_k)\) to \(f(SPD[i,j])\) for all node-pairs \((i,j)\) in a graph \(\mathcal{G}\) in the family of graphs indexed by \(\mathbb{G}_n\), and we define \(\phi_1\) to be that piecewise-linear map. The detailed analysis can be found in Appendix \ref{asymproof}).

Proposition \ref{proposition1} states that the attention mechanism defined in \ref{alpha} can capture (i) shortest path distances between all pairs of nodes and (ii) all weighted combinations of powers of the normalized adjacency matrix \(A_{norm}\) allowing us to perform aggregations over any \(k-\)hop neighborhood where \(k \in \mathbb{Z}\). 
The function \(\phi_2\) quantifies the "importance" of the frequency \(\lambda_k\) in the sum in \ref{alpha}. Traditional distance metrics like the biharmonic distance \citep{BiharmonicDistance} or the diffusion distance \citep{DiffusionMaps, GeometricDL}, weigh smaller frequencies more heavily, assuming that higher frequencies play a smaller role in the measurement of relative distance between nodes. We make no such assumption, rather allowing the dependence on the frequency to be learnt from data by parameterizing the function \(\phi_2\) appropriately. 

We now show that the proposed attention mechanism does not suffer from eigenvector sign/basis-invariance issues, making prior interventions like SignNet/BasisNet \citep{SignNet} redundant.

\begin{theorem} \label{proposition2}
    The attention matrix \(\alpha\) computed using equations \ref{sigma} and \ref{alpha}, is invariant to the choice of the sign of the eigenvectors \(u_k\), and in the general case, to the choice of basis of the eigenspace corresponding to the eigenvalue \(\lambda_k\) having geometric multiplicity > 1. 
\end{theorem}

\textit{Proof:} The sign-invariance can be easily inferred by noting that eigenvectors \(u_k\) appear in the computation of \(\alpha\) as multiplied pairs of terms 
(\ref{sigma}). The invariance to basis is due to the fact that in \ref{alpha}, the eigenvectors \(U_k = [u_{k_1}, ..., u_{k_{d_i}}] \in \mathbb{R}^{n\times d_i}\) corresponding to the same eigenvalue \(\lambda_k\) occur together as terms of the form \(h(U_kU_k^T)\), for some continuous function \(h\). Since any other orthonormal basis is of the form \(V_k = U_kQ\) for some orthogonal \(Q \in O(d_i) \subseteq \mathbb{R}^{d_i\times d_i}\), we have \(h(V_kV_k^T) = h(U_kU_k^T)\), implying that the choice of basis does not affect the computation in \ref{alpha}.



\begin{figure}
\centering
\begin{subfigure}{.5\textwidth}
  \centering
  \includegraphics[]{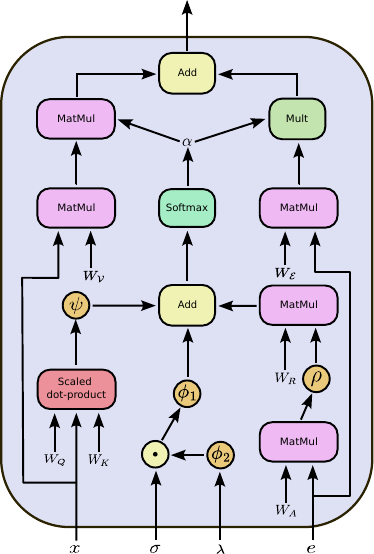}
  \caption{SAA: Spectrum-aware attention}
  \label{fig:propagation}
\end{subfigure}%
\begin{subfigure}{.5\textwidth}
  \centering
  \includegraphics[]{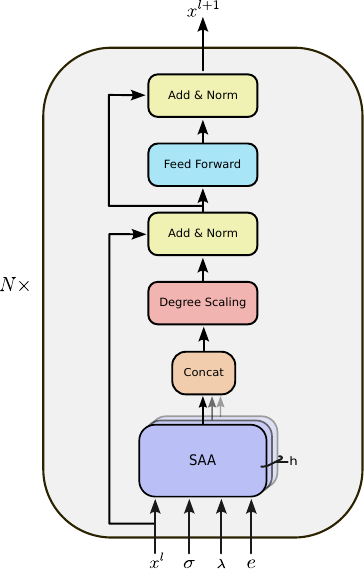}
  \caption{A layer of \textsc{Eigenformer}}
  \label{fig:layer}
\end{subfigure}
\caption{\textsc{Eigenformer} Architecture}
\label{fig:eigenformer}
\end{figure}

\subsection{\textsc{Eigenformer} architecture}
For a layer \(l\) of \textsc{Eigenformer}, input node features \(x_i^l \in \mathbb{R}^{d_\mathcal{V}}\) for \(i=1,...,\vert \mathcal{V} \vert\) and input edge features \(e_{ij} \in \mathbb{R}^{d_\mathcal{E}}\) for \(i,j=1,...,\vert \mathcal{V}\vert\), the node feature propagation is performed as follows:

\begin{equation}\label{alpha_spec}
{\alpha^{lh}_{spec}} = \phi_{1}^{lh}\Bigl(\sum_{k=1}^{\vert\mathcal{V}\vert}\sigma_k[i,j]\phi_{2}^{lh}(\lambda_k)\Bigl)
\end{equation}

\begin{equation}
{\alpha^{lh}_{feat}} = \psi\Bigl(\frac{(W^{lh}_Qx_i^l)^T(W^{lh}_Kx_j^l)}{\sqrt{d/H}}\Bigl) + W^{lh}_R\rho\bigl(W^{lh}_Ae_{ij}\bigl)
\end{equation}

\begin{equation}
    \alpha^{lh}[i,j] = softmax_{j \in \mathcal{V}}({\alpha^{lh}_{spec}} + {\alpha^{lh}_{feat}})
\end{equation}

\begin{equation}
    \hat{x}_i^{l+1} = W^l_\mathcal{O} \concat_{h=1}^{H} \sum_{j \in \mathcal{V}} \alpha^{lh}[i,j]\cdot \bigl(W^{lh}_\mathcal{V}x_j^l + W^{lh}_\mathcal{E}e_{ij}\bigl)
\end{equation}

\begin{equation}
    \Tilde{x}_i^{l+1} = \hat{x}_i^{l+1} \odot \theta_1 + log(1+d_i) \cdot \hat{x}_i^{l+1} \odot \theta_2
\end{equation}

\begin{equation}
    x_i^{\prime l+1} = BN(x_i^l + \Tilde{x}_i^{l+1})
\end{equation}

\begin{equation}
    x_i^{l+1} = BN(x_i^{\prime l+1} + FFN(x_i^{\prime l+1}))
\end{equation}

where \(H\) is the number of attention heads, \(\concat\) denotes concatenation, \(d\) is the hidden dimension, \(d_i\) are node degrees, \(W^l_{h,\mathcal{V}} \in \mathbb{R}^{\frac{d}{H}\times d_\mathcal{V}}\), \(W^l_{h,\mathcal{E}} \in \mathbb{R}^{\frac{d}{H}\times d_\mathcal{E}}\), \(W^l_\mathcal{O} \in \mathbb{R}^{d \times d}\), \(\theta_1 \in \mathbb{R}^d\) and \(\theta_2 \in \mathbb{R}^d\) are learnable weights (biases omitted for clarity), \(BN\) denotes Batch Normalization while \(FFN\) denotes a feed-forward network. Further, \(\rho\) is the ReLU activation function, \(\psi\) is the (optional) signed-square-root function: \(\psi(x) = ReLU(x)^\frac{1}{2} - ReLU(-x)^\frac{1}{2}\) which helps in stabilizing training as reported in \citet{GRIT}. We set \(\phi_1^{lh}\) and \(\phi_2^{lh}\) to be two-layer MLPs in all our experiments. 

\(\alpha_{feat}\) is an optional attention component that we introduce to study the gains that node and edge feature attention brings over solitary spectral attention \(\alpha_{spec}\). We do not include edge-feature propagation due to the added complexity and marginal benefits in performance. Notably however, we found degree-scaling followed by batch-normalization to be highly effective in incorporating node-degree information. The final architecture of a layer of \textsc{Eigenformer} is illustrated in Figure \ref{fig:layer} along with the spectrum-aware attention mechanism that it uses in Figure \ref{fig:propagation}.


\begin{table}[t!]
\small
  \caption{Test performance on five benchmarks from \citet{Benchmarking}. Due to limited training resources, we show results for only a single training run while other baselines report mean \(\pm\) standard deviation of 4 runs with different random seeds. Color codes denote top \textcolor{Green}{first}, \textcolor{Peach}{second}, \textcolor{Plum}{third} results.}
  \label{gnn-benchmark-table}
  \centering
  \begin{tabular}{lccccc}
    \toprule
    Model    & ZINC & MNIST & CIFAR10 & PATTERN & CLUSTER\\
    \cmidrule(l){2-6}
             & MAE \(\downarrow\)  & Accuracy\(\uparrow\) & Accuracy\(\uparrow\) & Wt. Acc.\(\uparrow\) & Accuracy\(\uparrow\) \\
    \midrule
    GCN & 0.367\(\pm\).011 & 90.705\(\pm\).218 & 55.710\(\pm\).381 & 71.892\(\pm\).334 & 68.498\(\pm\).976\\
    GIN & 0.526\(\pm\).051 & 96.485\(\pm\).252 & 55.255\(\pm\)1.527 & 85.387\(\pm\).136 & 64.716\(\pm\)1.553\\
    GAT & 0.384\(\pm\).007 & 95.535\(\pm\).205 & 64.223\(\pm\).455 & 78.271\(\pm\).186 & 70.587\(\pm\).447\\
    GatedGCN & 0.282\(\pm\).015 & 97.340\(\pm\).143 & 67.312\(\pm\).311 & 85.568\(\pm\).088 & 73.840\(\pm\).326\\
    GatedGCN-LSPE & 0.090\(\pm\).001 & - & - & - & -\\
    PNA & 0.188\(\pm\).004 & 97.940\(\pm\).12 & 70.35\(\pm\).63 & - & -\\
    DGN & 0.168\(\pm\).003 & - & \textcolor{Plum}{72.838\(\pm\).417} & 86.680\(\pm\).034 & -\\
    GSN & 0.101\(\pm\).010 & - & - & - & -\\
    \midrule
    CIN & 0.079\(\pm\).006 & - & - & - & -\\
    CRaW1 & 0.085\(\pm\).004 & 97.944\(\pm\).050 & 69.013\(\pm\).259 & - & -\\
    GIN-AK+ & 0.080\(\pm\).001 & - & 72.19\(\pm\).13 & \textcolor{Peach}{86.850\(\pm\).057} & -\\
    \midrule
    SAN & 0.139\(\pm\).006 & - & - & 86.581\(\pm\).037 & 76.691\(\pm\).65\\
    Graphormer & 0.122\(\pm\).006 & - & - & - & -\\
    K-Subgraph SAT & 0.094\(\pm\).008 & - & - & \textcolor{Plum}{86.848\(\pm\).037} & 77.856\(\pm\).104\\
    EGT & 0.108\(\pm\).009 & \textcolor{Plum}{98.173\(\pm\).087} & 68.702\(\pm\).409 & 86.821\(\pm\).020 & \textcolor{Peach}{79.232\(\pm\).348}\\
    Graphormer-URPE & 0.086\(\pm\).007 & - & - & - & -\\
    Graphormer-GD & 0.081\(\pm\).009 & - & - &  - & -\\
    GPS & \textcolor{Peach}{0.070\(\pm\).004} & 98.051\(\pm\).126 & 72.298\(\pm\).356 &  86.685\(\pm\).059 & 78.016\(\pm\).180\\
    GRIT & \textcolor{Green}{0.059\(\pm\).002} & 98.108\(\pm\).111 & \textcolor{Green}{76.468\(\pm\).881} &  \textcolor{Green}{87.196\(\pm\).076} & \textcolor{Green}{80.026\(\pm\).277}\\
    EXPHORMER & - & \textcolor{Green}{98.550\(\pm\).039} & \textcolor{Peach}{74.69\(\pm\).125} &  86.740\(\pm\).015 & \textcolor{Plum}{78.070\(\pm\).037}\\
    \midrule
    \textsc{Eigenformer} & \textcolor{Plum}{0.077} & 97.142 & 67.316 & 86.680 & 77.262\\
    \textsc{Eigenformer} (feat.) & 0.089 & \textcolor{Peach}{98.362} & 70.194  & 86.738 & 77.456 \\
    \bottomrule
  \end{tabular}
\end{table}

\begin{table}[t!]
  \caption{Test performance on a benchmark from the long-range graph benchmarks (LRGB) work \citep{LRGB}. Due to limited training resources, we show results for only a single training run while other baselines report mean \(\pm\) standard deviation of 4 runs with different random seeds. Color codes denote top \textcolor{Green}{first}, \textcolor{Peach}{second}, \textcolor{Plum}{third} results.}
  \label{lrgb-table}
  \centering
  \begin{tabular}{lcc}
    \toprule 
    Model    & Peptides-func & Peptides-struct\\
    \cmidrule(l){2-3}
             & AP \(\uparrow\)  & MAE\(\downarrow\)\\
    \midrule
    GCN & 0.5930\(\pm\)0.0023 & 0.3496\(\pm\)0.0013\\
    GINE & 0.5498\(\pm\)0.0079 & 0.3547\(\pm\)0.0045\\
    GatedGCN & 0.5864\(\pm\)0.0035 & 0.3420\(\pm\)0.0013\\
    GatedGCN+RWSE & 0.6069\(\pm\)0.0035 & 0.3357\(\pm\)0.0006\\
    \midrule
    Transformer+LapPE & 0.6326\(\pm\)0.0126 & \textcolor{Plum}{0.2529\(\pm\)0.0016}\\
    SAN+LapPE & 0.6384\(\pm\)0.0121 & 0.2683\(\pm\)0.0043\\
    SAN+RWSE & 0.6439\(\pm\)0.0075 & 0.2545\(\pm\)0.0012\\
    GPS & \textcolor{Peach}{0.6535\(\pm\)0.0041} & 0.2500\(\pm\)0.0012\\
    GRIT & \textcolor{Green}{0.6988\(\pm\)0.0082} & \textcolor{Green}{0.2460\(\pm\)0.0012}\\
    EXPHORMER & \textcolor{Plum}{0.6527\(\pm\)0.0043} & \textcolor{Peach}{0.2481\(\pm\)0.0007}\\
    \midrule
    \textsc{Eigenformer} & 0.6414 & 0.2599 \\
    \textsc{Eigenformer} (feat.) & 0.6317 & 0.2687\\
    \bottomrule
  \end{tabular}
\end{table}
\raggedbottom

\begin{figure}[b!]
\centering
\begin{subfigure}{.329\textwidth}
  \centering
  \includegraphics[]{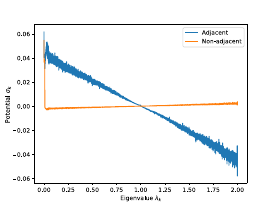}
  \caption{ZINC}
  \label{fig:zinc}
\end{subfigure}
\begin{subfigure}{.329\textwidth}
  \centering
  \includegraphics[]{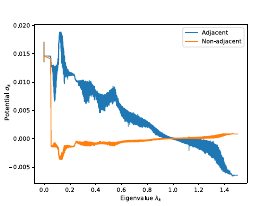}
  \caption{MNIST}
  \label{fig:mnist}
\end{subfigure}
\begin{subfigure}{.329\textwidth}
  \centering
  \includegraphics[]{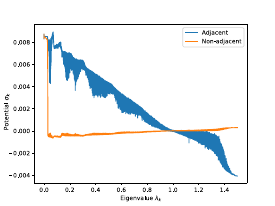}
  \caption{CIFAR10}
  \label{fig:cifar}
\end{subfigure}
\begin{subfigure}{.329\textwidth}
  \centering
  \includegraphics[]{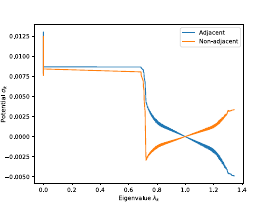}
  \caption{PATTERN}
  \label{fig:pattern}
\end{subfigure}
\begin{subfigure}{.329\textwidth}
  \centering
  \includegraphics[]{images/pattern_small_2.pdf}
  \caption{CLUSTER}
  \label{fig:cluster}
\end{subfigure}
\begin{subfigure}{.329\textwidth}
  \centering
  \includegraphics[]{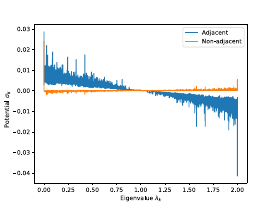}
  \caption{Peptides-func/struct}
  \label{fig:peptides-func}
\end{subfigure}
\caption{(Smoothed) potential \(\sigma_k\) vs eigenvalue \(\lambda_k\)} for adjacent and non-adjacent nodes
\label{fig:rel-sim}
\end{figure}

\section{Experimental Results}

We evaluate \textsc{Eigenformer} on five benchmarks from the widely used Benchmarking GNNs work \citep{Benchmarking}, viz. ZINC, MNIST, CIFAR10, PATTERN and CLUSTER. These datasets include different graph learning objectives like node classification, graph classification and graph regression. They also cover widely different domains ranging from chemistry and computer vision to synthetically generated graphs. Further, we report numbers on two benchmarks from the Long-Range Graph Benchmark (LRGB) \citep{LRGB} to gauge how well SAA does in capturing long-range dependencies. Further details about the experimental setup can be found in appendix \ref{expdetails}.


\textbf{Baselines:} We compare our method to a variety of SOTA GNN methods including (hybrid) Graph Transformers: GraphGPS \citet{GraphGPS}, GRIT \citet{GRIT}, EGT \citet{Hussain}, SAN \citet{SAN}, Graphormer \citet{Ying}, K-Subgraph SAT \citet{Chen2022StructureAwareTF} and EXPHORMER \citet{Exphormer}, MP-GNNs: GCN \citet{GCN}, GAT \citet{GAT}, GIN \citet{xu2018how}, PNA \citet{PNA}, GatedGCN \citet{GatedGCN}, DGN \citet{DGN}, GSN \citet{GSN} and other performant models like CIN \citet{CIN}, CRaW1 \citet{CRaW1} and GIN-AK+ \citet{GinAK}.

\textbf{Benchmarking GNNs \citep{Benchmarking}:} Table \ref{gnn-benchmark-table} compares different methods under the parameter budget used in \citet{GraphGPS} (\(\sim\)500K parameters for ZINC, PATTERN and CLUSTER, \(\sim\)100K parameters for MNIST and CIFAR10). \textsc{Eigenformer} routinely beats all the MP-GNNs and performs comparably to popular Graph Transformers that use different PEs like SAN, Subgraph-SAT, Graphormer, GRIT etc. Further, \textsc{Eigenformer} with both spectral and feature attention does marginally better for these datasets except on ZINC, where spectral attention seems to be adequate. Feature attention appears to be most beneficial for the two directed-graph datasets: MNIST and CIFAR10.

\textbf{Long-range graph benchmark \citep{LRGB}:} 
We also report numbers on two benchmarks from the LRGB work \citep{LRGB} viz. Peptides-func (multilabel graph classification) and Peptides-struct (graph regression) in Table \ref{lrgb-table}. Again, \textsc{Eigenformer} beats all the MP-GNN models comfortably and is competitive with other Graph Transformers albeit falling short of featuring among the best performers. Interestingly, feature attention is detrimental to performance for these long-range interactions. In conjunction with the results from Table \ref{gnn-benchmark-table}, our experiments suggest that feature information is not very helpful in determining how messages ought to be exchanged between nodes.

\textbf{Visualization of node potentials:} It is instructive to visualize the difference in the distribution of potentials (as defined in \ref{sigma}) between adjacent and non-adjacent nodes. Figure \ref{fig:rel-sim} shows the (smoothed, average) potential distribution w.r.t frequency for all the datasets used in our work. In general, adjacent nodes have higher potentials at the lower end of the spectrum that vary inversely with frequency. Non-adjacent nodes on the other hand, have smaller potentials which increase towards the higher end of the spectrum. We posit that these distributional differences allow the attention mechanism to learn to discern local interactions from long-range interactions.

\begin{figure}
\centering
\begin{subfigure}{.329\textwidth}
  \centering
  \includegraphics[]{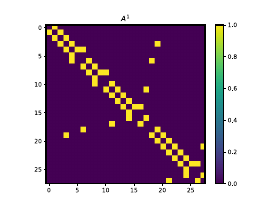}
  \caption{}
  \label{fig:A1}
\end{subfigure}
\begin{subfigure}{.329\textwidth}
  \centering
  \includegraphics[]{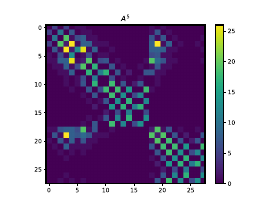}
  \caption{}
  \label{fig:A5}
\end{subfigure}
\begin{subfigure}{.329\textwidth}
  \centering
  \includegraphics[]{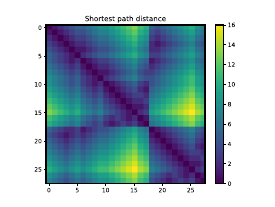}
  \caption{}
  \label{fig:spd}
\end{subfigure}
\begin{subfigure}{.329\textwidth}
  \centering
  \includegraphics[]{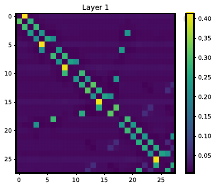}
  \caption{}
  \label{fig:Aatt}
\end{subfigure}
\begin{subfigure}{.329\textwidth}
  \centering
  \includegraphics[]{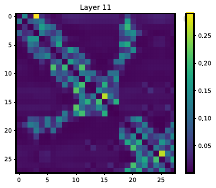}
  \caption{}
  \label{fig:A5att}
\end{subfigure}
\begin{subfigure}{.329\textwidth}
  \centering
  \includegraphics[]{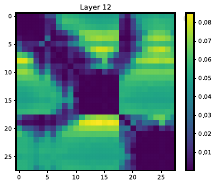}
  \caption{}
  \label{fig:spdatt}
\end{subfigure}
\caption{k-hop neighborhood, shortest-path distance and learned attention matrices for an example graph from the ZINC dataset}
\label{fig:adj-att}
\end{figure}

\textbf{Visualization of learned attention matrices:}
Figure \ref{fig:adj-att} shows the adjacency matrices (1-hop: \ref{fig:A1}, 5-hop: \ref{fig:A5}) and the shortest-path distance matrix (\ref{fig:spd}) for an example graph from the ZINC dataset, along with visualizations of learned attention patterns from the trained model that most closely mimic the corresponding neighborhood pattern (\ref{fig:Aatt}, \ref{fig:A5att}, \ref{fig:spdatt}).
For example, Figure \ref{fig:A5att} shows the attention pattern of a head from Layer 11 of the learned model, which seemingly tries to reproduce the 5-hop neighborhood interactions similar to those shown in Figure \ref{fig:A5}, providing empirical evidence that \textsc{Eigenformer} can learn k-hop interactions.



\textbf{Limitations:} A couple of open challenges remain however, including the final test performance and computational complexity of \textsc{Eigenformer}. With regards to performance, we believe a more extensive hyperparameter search coupled with edge-propagation might aid in closing the gap to the current SOTA in Graph Transformers. The main bottleneck w.r.t complexity is equation \ref{alpha_spec} which can be implemented using in-place operations with \(O(N^2)\) memory but in \(O(N^3)\) time. The payoff for the added complexity is increased representational capacity, as SAA can learn \(k\)-hop interactions for \textit{any} \(k \in \mathbb{Z}\) as against upper-bounded \(k\)-hop neighborhoods in previous works. An obvious way to reduce the time complexity to \(O(N^2k)\) is by using a subset of the frequencies. We test this intervention on the ZINC dataset, by randomly choosing \(k\) out of \(N\) eigenvalues for each example in the data at each training step. Table \ref{N2k-table} and Figure \ref{N2k-fig} show the change in final test performance with varying \(k\). The test performance remains impressive even when using \(\sim40\%\) of the eigenvalues, suggesting a graceful degradation in performance with decreasing number of eigenvalues (\(k\)) used.


    \begin{minipage}{0.49\textwidth}
        \centering
        \captionof{table}{Test MAE for the ZINC dataset for varying number (k) of eigenvalues used. N=37.}
          \begin{tabular}[b]{ccc}    
            \toprule
            k & k/N & Test MAE \(\downarrow\) \\
            \midrule
            37 & 1.0 & 0.089\\
            30 & 0.81 & 0.097\\
            22 & 0.59 & 0.108\\
            15 & 0.41 & 0.133\\
            8 & 0.22 & 0.180\\
            \bottomrule
          \end{tabular}
      \label{N2k-table}
    \end{minipage}
    \hfill
    \begin{minipage}{0.49\textwidth}
        \centering
        \includegraphics[]{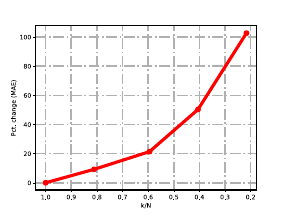}
        \captionof{figure}{Percentage change in MAE with decreasing number (k) of eigenvalues used}
        \label{N2k-fig}
    \end{minipage}

\section{Related Work}
Considering the importance of graph connectivity/topology information for any GNN architecture, there have been many attempts to incorporate graph inductive biases in Transformers following the introduction of Graph Transformers in \citet{li2019graph}.

\textbf{Graph inductive biases in Transformers:} \citet{DwivediBresson} performs attention only over the 1-hop neighborhood of nodes in addition to the use of Laplacian eigenvectors as node PEs. With Graph-BERT, \citet{Zhang2020GraphBertOA} proposes to train Transformers using sampled linkless (no edges) subgraphs within their local contexts, designed with an emphasis on parallelization and mainly as a pretraining technique. They use a number of PEs like the WL Absolute Role Embedding, Intimacy-based Relative Positional Embedding and Hop-based Relative Distance Embedding, in order to retain the original graph structural information in the linkless subgraphs. 

\citet{SAN} uses a Transformer encoder to create learnable PEs from a combination of eigenvalues and eigenvectors which are then fed to the main Transformer network. \citet{RWPE} decouples the node features and the node PEs with a separate propagation step for the PEs. They consider two different initializations for the PEs: Laplacian PE (LapPE) and Random Walk PE (RWPE), and show that RWPE outperforms LapPE, indicating that learning the sign invariance ambiguity in LapPE is more difficult than not exactly having unique node representation for each node. \citet{Ying} proposes the use of a fixed centrality encoding to inject degree information in their Graphormer.

With GraphiT, \citet{Mialon2021GraphiTEG} enumerates, encodes and aggregates local sub-structures such as paths of length-k using Graph convolutional kernel networks (GCKN) \citep{GCKN} to create node PEs. \citet{GRIT} proposes GRIT, again using PEs initialized with random walk probabilities but makes them flexible and learnable by composing with an MLP, and provide theoretical proof of the expressivity of such a composition. Further, they inject degree information using an adaptive degree-scaler \citep{PNA} in every layer, noting that attention mechanisms are invariant to node-degrees.

\textbf{Structure-aware attention mechanisms:} More similar to our work, there has also been research on creating structure-aware attention mechanisms in order to infuse node structural and neighborhood similarity information directly in the attention logic. \citet{Mialon2021GraphiTEG} views self-attention as kernel smoothing and bias the attention scores using diffusion or k-step random-walk kernels. \citet{Chen2022StructureAwareTF} argues that such a modification is deficient in filtering out structurally dissimilar nodes, and consider a more generalized kernel accounting for local substructures like k-subtrees and k-subgraphs, around each node. In a similar approach, \citet{g2023satg} aggregates node feature information from k-subtrees but augment their Transformer with a global node and train positional encodings using the Neighbourhood Contrastive Loss (NCE). \citet{Ying} proposed to inject spatial and edge information by adding learnable encodings to the attention scores before normalization, while \citep{menegaux2023selfattention} augments that idea further by learning a different attention score per hidden dimension in their so-called Chromatic Self-Attention (CSA). Learned edge encodings are also incorporated in \citet{Hussain} both as a bias to the attention score before normalization and as a sigmoid-gate after normalization.

In contrast to previous work, to the best of our knowledge, there have been no attempts to factorize the attention matrix solely into fixed spectral potentials and learnable frequency importances.

\section{Conclusion}
Observing the wide variety of positional encoding schemes for Graph Transformers in the literature, we asked the question whether such encodings are essential to ace graph representational learning objectives. We introduced a novel spectrum-aware attention (SAA) mechanism that incorporates structural graph inductive biases while remaining flexible enough to learn from features in the data. SAA was also shown to be invariant to sign or basis invariance issues that plagued previous PE design strategies. We proposed \textsc{Eigenformer}, a Graph Transformer with SAA at its core, that can be used for a variety of graph learning objectives with different task-specific heads. We empirically test the feasibility of our method through experiments on a number of common GNN benchmarks and find that it performs comparably to many popular GNN architectures employing different schemes for PE generation. We hope this work provides the impetus to the broader graph representational learning research community to pursue the development of other GNN architectures without the use of PEs.




{
\small

\bibliography{references}

\begin{thebibliography}{46}
\providecommand{\natexlab}[1]{#1}
\providecommand{\url}[1]{\texttt{#1}}
\expandafter\ifx\csname urlstyle\endcsname\relax
  \providecommand{\doi}[1]{doi: #1}\else
  \providecommand{\doi}{doi: \begingroup \urlstyle{rm}\Url}\fi

\bibitem[Veličković(2023)]{EverythingConnected}
Petar Veličković.
\newblock Everything is connected: Graph neural networks.
\newblock \emph{Current Opinion in Structural Biology}, 79:\penalty0 102538, 2023.
\newblock ISSN 0959-440X.
\newblock \doi{https://doi.org/10.1016/j.sbi.2023.102538}.
\newblock URL \url{https://www.sciencedirect.com/science/article/pii/S0959440X2300012X}.

\bibitem[Bruna et~al.(2014)Bruna, Zaremba, Szlam, and LeCun]{Bruna}
Joan Bruna, Wojciech Zaremba, Arthur Szlam, and Yann LeCun.
\newblock Spectral networks and locally connected networks on graphs.
\newblock In Yoshua Bengio and Yann LeCun, editors, \emph{2nd International Conference on Learning Representations, {ICLR} 2014, Banff, AB, Canada, April 14-16, 2014, Conference Track Proceedings}, 2014.
\newblock URL \url{http://arxiv.org/abs/1312.6203}.

\bibitem[Gilmer et~al.(2017)Gilmer, Schoenholz, Riley, Vinyals, and Dahl]{Quantum}
Justin Gilmer, Samuel~S. Schoenholz, Patrick~F. Riley, Oriol Vinyals, and George~E. Dahl.
\newblock Neural message passing for quantum chemistry.
\newblock In \emph{Proceedings of the 34th International Conference on Machine Learning - Volume 70}, ICML'17, page 1263–1272. JMLR.org, 2017.

\bibitem[Vaswani et~al.(2017)Vaswani, Shazeer, Parmar, Uszkoreit, Jones, Gomez, Kaiser, and Polosukhin]{Vaswani}
Ashish Vaswani, Noam Shazeer, Niki Parmar, Jakob Uszkoreit, Llion Jones, Aidan~N. Gomez, Lukasz Kaiser, and Illia Polosukhin.
\newblock Attention is all you need.
\newblock \emph{CoRR}, abs/1706.03762, 2017.
\newblock URL \url{http://arxiv.org/abs/1706.03762}.

\bibitem[Dwivedi and Bresson(2020)]{DwivediBresson}
Vijay~Prakash Dwivedi and Xavier Bresson.
\newblock A generalization of transformer networks to graphs.
\newblock \emph{CoRR}, abs/2012.09699, 2020.
\newblock URL \url{https://arxiv.org/abs/2012.09699}.

\bibitem[Kreuzer et~al.(2021)Kreuzer, Beaini, Hamilton, L{\'{e}}tourneau, and Tossou]{SAN}
Devin Kreuzer, Dominique Beaini, William~L. Hamilton, Vincent L{\'{e}}tourneau, and Prudencio Tossou.
\newblock Rethinking graph transformers with spectral attention.
\newblock \emph{CoRR}, abs/2106.03893, 2021.
\newblock URL \url{https://arxiv.org/abs/2106.03893}.

\bibitem[Chen et~al.(2022{\natexlab{a}})Chen, O’Bray, and Borgwardt]{StructureAware}
Dexiong Chen, Leslie O’Bray, and Karsten~M. Borgwardt.
\newblock Structure-aware transformer for graph representation learning.
\newblock In \emph{International Conference on Machine Learning}, 2022{\natexlab{a}}.
\newblock URL \url{https://api.semanticscholar.org/CorpusID:246634635}.

\bibitem[Ying et~al.(2021)Ying, Cai, Luo, Zheng, Ke, He, Shen, and Liu]{Ying}
Chengxuan Ying, Tianle Cai, Shengjie Luo, Shuxin Zheng, Guolin Ke, Di~He, Yanming Shen, and Tie-Yan Liu.
\newblock Do transformers really perform badly for graph representation?
\newblock In A.~Beygelzimer, Y.~Dauphin, P.~Liang, and J.~Wortman Vaughan, editors, \emph{Advances in Neural Information Processing Systems}, 2021.
\newblock URL \url{https://openreview.net/forum?id=OeWooOxFwDa}.

\bibitem[Ramp\'{a}\v{s}ek et~al.(2022)Ramp\'{a}\v{s}ek, Galkin, Dwivedi, Luu, Wolf, and Beaini]{GraphGPS}
Ladislav Ramp\'{a}\v{s}ek, Michael Galkin, Vijay~Prakash Dwivedi, Anh~Tuan Luu, Guy Wolf, and Dominique Beaini.
\newblock Recipe for a general, powerful, scalable graph transformer.
\newblock In S.~Koyejo, S.~Mohamed, A.~Agarwal, D.~Belgrave, K.~Cho, and A.~Oh, editors, \emph{Advances in Neural Information Processing Systems}, volume~35, pages 14501--14515. Curran Associates, Inc., 2022.
\newblock URL \url{https://proceedings.neurips.cc/paper_files/paper/2022/file/5d4834a159f1547b267a05a4e2b7cf5e-Paper-Conference.pdf}.

\bibitem[Hussain et~al.(2022)Hussain, Zaki, and Subramanian]{Hussain}
Md~Shamim Hussain, Mohammed~J. Zaki, and Dharmashankar Subramanian.
\newblock Global self-attention as a replacement for graph convolution.
\newblock In \emph{Proceedings of the 28th ACM SIGKDD Conference on Knowledge Discovery and Data Mining}, KDD '22, page 655–665, New York, NY, USA, 2022. Association for Computing Machinery.
\newblock ISBN 9781450393850.
\newblock \doi{10.1145/3534678.3539296}.
\newblock URL \url{https://doi.org/10.1145/3534678.3539296}.

\bibitem[Zhang et~al.(2023)Zhang, Luo, Wang, and He]{zhang2023rethinking}
Bohang Zhang, Shengjie Luo, Liwei Wang, and Di~He.
\newblock Rethinking the expressive power of {GNN}s via graph biconnectivity.
\newblock In \emph{The Eleventh International Conference on Learning Representations}, 2023.
\newblock URL \url{https://openreview.net/forum?id=r9hNv76KoT3}.

\bibitem[Alon and Yahav(2021)]{alon2021on}
Uri Alon and Eran Yahav.
\newblock On the bottleneck of graph neural networks and its practical implications.
\newblock In \emph{International Conference on Learning Representations}, 2021.
\newblock URL \url{https://openreview.net/forum?id=i80OPhOCVH2}.

\bibitem[Topping et~al.(2022)Topping, Giovanni, Chamberlain, Dong, and Bronstein]{topping2022understanding}
Jake Topping, Francesco~Di Giovanni, Benjamin~Paul Chamberlain, Xiaowen Dong, and Michael~M. Bronstein.
\newblock Understanding over-squashing and bottlenecks on graphs via curvature.
\newblock In \emph{International Conference on Learning Representations}, 2022.
\newblock URL \url{https://openreview.net/forum?id=7UmjRGzp-A}.

\bibitem[Li et~al.(2018)Li, Han, and Wu]{deeper}
Qimai Li, Zhichao Han, and Xiao-Ming Wu.
\newblock Deeper insights into graph convolutional networks for semi-supervised learning.
\newblock In \emph{Proceedings of the Thirty-Second AAAI Conference on Artificial Intelligence and Thirtieth Innovative Applications of Artificial Intelligence Conference and Eighth AAAI Symposium on Educational Advances in Artificial Intelligence}, AAAI'18/IAAI'18/EAAI'18. AAAI Press, 2018.
\newblock ISBN 978-1-57735-800-8.

\bibitem[Oono and Suzuki(2020)]{Oono2020Graph}
Kenta Oono and Taiji Suzuki.
\newblock Graph neural networks exponentially lose expressive power for node classification.
\newblock In \emph{International Conference on Learning Representations}, 2020.
\newblock URL \url{https://openreview.net/forum?id=S1ldO2EFPr}.

\bibitem[Xu et~al.(2019)Xu, Hu, Leskovec, and Jegelka]{xu2018how}
Keyulu Xu, Weihua Hu, Jure Leskovec, and Stefanie Jegelka.
\newblock How powerful are graph neural networks?
\newblock In \emph{International Conference on Learning Representations}, 2019.
\newblock URL \url{https://openreview.net/forum?id=ryGs6iA5Km}.

\bibitem[Loukas(2020)]{Loukas2020What}
Andreas Loukas.
\newblock What graph neural networks cannot learn: depth vs width.
\newblock In \emph{International Conference on Learning Representations}, 2020.
\newblock URL \url{https://openreview.net/forum?id=B1l2bp4YwS}.

\bibitem[Morris et~al.(2019)Morris, Ritzert, Fey, Hamilton, Lenssen, Rattan, and Grohe]{WL}
Christopher Morris, Martin Ritzert, Matthias Fey, William~L. Hamilton, Jan~Eric Lenssen, Gaurav Rattan, and Martin Grohe.
\newblock Weisfeiler and leman go neural: higher-order graph neural networks.
\newblock In \emph{Proceedings of the Thirty-Third AAAI Conference on Artificial Intelligence and Thirty-First Innovative Applications of Artificial Intelligence Conference and Ninth AAAI Symposium on Educational Advances in Artificial Intelligence}, AAAI'19/IAAI'19/EAAI'19. AAAI Press, 2019.
\newblock ISBN 978-1-57735-809-1.
\newblock \doi{10.1609/aaai.v33i01.33014602}.
\newblock URL \url{https://doi.org/10.1609/aaai.v33i01.33014602}.

\bibitem[Lim et~al.(2023)Lim, Robinson, Zhao, Smidt, Sra, Maron, and Jegelka]{SignNet}
Derek Lim, Joshua~David Robinson, Lingxiao Zhao, Tess Smidt, Suvrit Sra, Haggai Maron, and Stefanie Jegelka.
\newblock Sign and basis invariant networks for spectral graph representation learning.
\newblock In \emph{The Eleventh International Conference on Learning Representations}, 2023.
\newblock URL \url{https://openreview.net/forum?id=Q-UHqMorzil}.

\bibitem[Dwivedi et~al.(2022{\natexlab{a}})Dwivedi, Luu, Laurent, Bengio, and Bresson]{RWPE}
Vijay~Prakash Dwivedi, Anh~Tuan Luu, Thomas Laurent, Yoshua Bengio, and Xavier Bresson.
\newblock Graph neural networks with learnable structural and positional representations.
\newblock In \emph{International Conference on Learning Representations}, 2022{\natexlab{a}}.
\newblock URL \url{https://openreview.net/forum?id=wTTjnvGphYj}.

\bibitem[Ma et~al.(2023)Ma, Lin, Lim, Romero-Soriano, Dokania, Coates, Torr, and Lim]{GRIT}
Liheng Ma, Chen Lin, Derek Lim, Adriana Romero-Soriano, Puneet~K. Dokania, Mark Coates, Philip~H.S. Torr, and Ser-Nam Lim.
\newblock Graph inductive biases in transformers without message passing.
\newblock In \emph{Proceedings of the 40th International Conference on Machine Learning}, ICML'23. JMLR.org, 2023.

\bibitem[Hu et~al.()Hu, Fey, Ren, Nakata, Dong, and Leskovec]{ogb_lsc}
Weihua Hu, Matthias Fey, Hongyu Ren, Maho Nakata, Yuxiao Dong, and Jure Leskovec.
\newblock Ogb-lsc: A large-scale challenge for machine learning on graphs.
\newblock \emph{NeurIPS}, 34.
\newblock URL \url{https://par.nsf.gov/biblio/10396194}.

\bibitem[Chung and Yau(2000)]{Greens}
Fan Chung and S.-T. Yau.
\newblock Discrete green's functions.
\newblock \emph{Journal of Combinatorial Theory, Series A}, 91\penalty0 (1):\penalty0 191--214, 2000.
\newblock ISSN 0097-3165.
\newblock \doi{https://doi.org/10.1006/jcta.2000.3094}.
\newblock URL \url{https://www.sciencedirect.com/science/article/pii/S0097316500930942}.

\bibitem[Lipman et~al.(2010)Lipman, Rustamov, and Funkhouser]{BiharmonicDistance}
Yaron Lipman, Raif~M. Rustamov, and Thomas~A. Funkhouser.
\newblock Biharmonic distance.
\newblock \emph{ACM Trans. Graph.}, 29\penalty0 (3), jul 2010.
\newblock ISSN 0730-0301.
\newblock \doi{10.1145/1805964.1805971}.
\newblock URL \url{https://doi.org/10.1145/1805964.1805971}.

\bibitem[Coifman and Lafon(2006)]{DiffusionMaps}
Ronald~R. Coifman and Stéphane Lafon.
\newblock Diffusion maps.
\newblock \emph{Applied and Computational Harmonic Analysis}, 21\penalty0 (1):\penalty0 5--30, 2006.
\newblock ISSN 1063-5203.
\newblock \doi{https://doi.org/10.1016/j.acha.2006.04.006}.
\newblock URL \url{https://www.sciencedirect.com/science/article/pii/S1063520306000546}.
\newblock Special Issue: Diffusion Maps and Wavelets.

\bibitem[Bronstein et~al.(2017)Bronstein, Bruna, LeCun, Szlam, and Vandergheynst]{GeometricDL}
Michael~M. Bronstein, Joan Bruna, Yann LeCun, Arthur Szlam, and Pierre Vandergheynst.
\newblock Geometric deep learning: Going beyond euclidean data.
\newblock \emph{IEEE Signal Processing Magazine}, 34\penalty0 (4):\penalty0 18--42, 2017.
\newblock \doi{10.1109/MSP.2017.2693418}.

\bibitem[Dwivedi et~al.(2020)Dwivedi, Joshi, Laurent, Bengio, and Bresson]{Benchmarking}
Vijay~Prakash Dwivedi, Chaitanya~K. Joshi, Thomas Laurent, Yoshua Bengio, and Xavier Bresson.
\newblock Benchmarking graph neural networks.
\newblock \emph{CoRR}, abs/2003.00982, 2020.
\newblock URL \url{https://arxiv.org/abs/2003.00982}.

\bibitem[Dwivedi et~al.(2022{\natexlab{b}})Dwivedi, Ramp\'{a}\v{s}ek, Galkin, Parviz, Wolf, Luu, and Beaini]{LRGB}
Vijay~Prakash Dwivedi, Ladislav Ramp\'{a}\v{s}ek, Michael Galkin, Ali Parviz, Guy Wolf, Anh~Tuan Luu, and Dominique Beaini.
\newblock Long range graph benchmark.
\newblock In S.~Koyejo, S.~Mohamed, A.~Agarwal, D.~Belgrave, K.~Cho, and A.~Oh, editors, \emph{Advances in Neural Information Processing Systems}, volume~35, pages 22326--22340. Curran Associates, Inc., 2022{\natexlab{b}}.
\newblock URL \url{https://proceedings.neurips.cc/paper_files/paper/2022/file/8c3c666820ea055a77726d66fc7d447f-Paper-Datasets_and_Benchmarks.pdf}.

\bibitem[Chen et~al.(2022{\natexlab{b}})Chen, O’Bray, and Borgwardt]{Chen2022StructureAwareTF}
Dexiong Chen, Leslie O’Bray, and Karsten~M. Borgwardt.
\newblock Structure-aware transformer for graph representation learning.
\newblock In \emph{International Conference on Machine Learning}, 2022{\natexlab{b}}.
\newblock URL \url{https://api.semanticscholar.org/CorpusID:246634635}.

\bibitem[Shirzad et~al.(2023)Shirzad, Velingker, Venkatachalam, Sutherland, and Sinop]{Exphormer}
Hamed Shirzad, Ameya Velingker, Balaji Venkatachalam, Danica~J. Sutherland, and Ali~Kemal Sinop.
\newblock Exphormer: sparse transformers for graphs.
\newblock In \emph{Proceedings of the 40th International Conference on Machine Learning}, ICML'23. JMLR.org, 2023.

\bibitem[Kipf and Welling(2017)]{GCN}
Thomas~N. Kipf and Max Welling.
\newblock Semi-supervised classification with graph convolutional networks.
\newblock In \emph{International Conference on Learning Representations}, 2017.
\newblock URL \url{https://openreview.net/forum?id=SJU4ayYgl}.

\bibitem[Veličković et~al.(2018)Veličković, Cucurull, Casanova, Romero, Liò, and Bengio]{GAT}
Petar Veličković, Guillem Cucurull, Arantxa Casanova, Adriana Romero, Pietro Liò, and Yoshua Bengio.
\newblock Graph attention networks.
\newblock In \emph{International Conference on Learning Representations}, 2018.
\newblock URL \url{https://openreview.net/forum?id=rJXMpikCZ}.

\bibitem[Corso et~al.(2020)Corso, Cavalleri, Beaini, Li\`{o}, and Velickovic]{PNA}
Gabriele Corso, Luca Cavalleri, Dominique Beaini, Pietro Li\`{o}, and Petar Velickovic.
\newblock Principal neighbourhood aggregation for graph nets.
\newblock In \emph{Proceedings of the 34th International Conference on Neural Information Processing Systems}, NIPS'20, Red Hook, NY, USA, 2020. Curran Associates Inc.
\newblock ISBN 9781713829546.

\bibitem[Bresson and Laurent(2018)]{GatedGCN}
Xavier Bresson and Thomas Laurent.
\newblock Residual gated graph convnets, 2018.
\newblock URL \url{https://openreview.net/forum?id=HyXBcYg0b}.

\bibitem[Beaini et~al.(2021)Beaini, Passaro, Letourneau, Hamilton, Corso, and Li{\`o}]{DGN}
Dominique Beaini, Saro Passaro, Vincent Letourneau, William~L. Hamilton, Gabriele Corso, and Pietro Li{\`o}.
\newblock Directional graph networks, 2021.
\newblock URL \url{https://openreview.net/forum?id=FUdBF49WRV1}.

\bibitem[Bouritsas et~al.(2023)Bouritsas, Frasca, Zafeiriou, and Bronstein]{GSN}
Giorgos Bouritsas, Fabrizio Frasca, Stefanos Zafeiriou, and Michael~M. Bronstein.
\newblock Improving graph neural network expressivity via subgraph isomorphism counting.
\newblock \emph{IEEE Transactions on Pattern Analysis and Machine Intelligence}, 45\penalty0 (1):\penalty0 657--668, 2023.
\newblock \doi{10.1109/TPAMI.2022.3154319}.

\bibitem[Bodnar et~al.(2021)Bodnar, Frasca, Wang, Otter, Mont{\'u}far, Lio’, and Bronstein]{CIN}
Cristian Bodnar, Fabrizio Frasca, Yu~Guang Wang, Nina Otter, Guido Mont{\'u}far, Pietro Lio’, and Michael~M. Bronstein.
\newblock Weisfeiler and lehman go topological: Message passing simplicial networks.
\newblock \emph{ArXiv}, abs/2103.03212, 2021.
\newblock URL \url{https://api.semanticscholar.org/CorpusID:232110693}.

\bibitem[T{\"o}nshoff et~al.(2021)T{\"o}nshoff, Ritzert, Wolf, and Grohe]{CRaW1}
Jan T{\"o}nshoff, Martin Ritzert, Hinrikus Wolf, and Martin Grohe.
\newblock Graph learning with 1d convolutions on random walks.
\newblock \emph{ArXiv}, abs/2102.08786, 2021.
\newblock URL \url{https://api.semanticscholar.org/CorpusID:263883878}.

\bibitem[Zhao et~al.(2021)Zhao, Jin, Akoglu, and Shah]{GinAK}
Lingxiao Zhao, Wei Jin, Leman Akoglu, and Neil Shah.
\newblock From stars to subgraphs: Uplifting any gnn with local structure awareness.
\newblock \emph{ArXiv}, abs/2110.03753, 2021.
\newblock URL \url{https://api.semanticscholar.org/CorpusID:238531375}.

\bibitem[Li et~al.(2019)Li, Liang, Hu, Chen, and Xing]{li2019graph}
Yuan Li, Xiaodan Liang, Zhiting Hu, Yinbo Chen, and Eric~P. Xing.
\newblock Graph transformer, 2019.
\newblock URL \url{https://openreview.net/forum?id=HJei-2RcK7}.

\bibitem[Zhang et~al.(2020)Zhang, Zhang, Sun, and Xia]{Zhang2020GraphBertOA}
Jiawei Zhang, Haopeng Zhang, Li~Sun, and Congying Xia.
\newblock Graph-bert: Only attention is needed for learning graph representations.
\newblock \emph{ArXiv}, abs/2001.05140, 2020.
\newblock URL \url{https://api.semanticscholar.org/CorpusID:210698881}.

\bibitem[Mialon et~al.(2021)Mialon, Chen, Selosse, and Mairal]{Mialon2021GraphiTEG}
Gr{\'e}goire Mialon, Dexiong Chen, Margot Selosse, and Julien Mairal.
\newblock Graphit: Encoding graph structure in transformers.
\newblock \emph{ArXiv}, abs/2106.05667, 2021.
\newblock URL \url{https://api.semanticscholar.org/CorpusID:235390675}.

\bibitem[Chen et~al.(2020)Chen, Jacob, and Mairal]{GCKN}
Dexiong Chen, Laurent Jacob, and Julien Mairal.
\newblock Convolutional kernel networks for graph-structured data.
\newblock In \emph{Proceedings of the 37th International Conference on Machine Learning}, ICML'20. JMLR.org, 2020.

\bibitem[G et~al.(2023)G, Patnala, Vasava, Sethi, and Gupta]{g2023satg}
Sumedh~B G, Sanjay Patnala, Himil Vasava, Akshay Sethi, and Sonia Gupta.
\newblock {SATG} : Structure aware transformers on graphs for node classification.
\newblock In \emph{NeurIPS 2023 Workshop: New Frontiers in Graph Learning}, 2023.
\newblock URL \url{https://openreview.net/forum?id=EVp40Cz0PR}.

\bibitem[Menegaux et~al.(2023)Menegaux, Jehanno, Selosse, and Mairal]{menegaux2023selfattention}
Romain Menegaux, Emmanuel Jehanno, Margot Selosse, and Julien Mairal.
\newblock Self-attention in colors: Another take on encoding graph structure in transformers.
\newblock \emph{Transactions on Machine Learning Research}, 2023.
\newblock ISSN 2835-8856.
\newblock URL \url{https://openreview.net/forum?id=3dQCNqqv2d}.

\bibitem[Hornik et~al.(1989)Hornik, Stinchcombe, and White]{HORNIK1989359}
Kurt Hornik, Maxwell Stinchcombe, and Halbert White.
\newblock Multilayer feedforward networks are universal approximators.
\newblock \emph{Neural Networks}, 2\penalty0 (5):\penalty0 359--366, 1989.
\newblock ISSN 0893-6080.
\newblock \doi{https://doi.org/10.1016/0893-6080(89)90020-8}.
\newblock URL \url{https://www.sciencedirect.com/science/article/pii/0893608089900208}.

\end{thebibliography}



}

\appendix
\section{Proofs}
\subsection{Proposition \ref{proposition1}}\label{asymproof}
(1) For an undirected graph \(\mathcal{G}=(\mathcal{V}, \mathcal{E})\) with adjacency matrix \(A\) and degree matrix \(D\), we have:

\begin{equation*}
    L_{norm} = I - D^{-\frac{1}{2}}AD^{-\frac{1}{2}} = I - A_{norm}
\end{equation*}

Since \(L_{norm}\) is real and symmetric, it is diagonalizable and admits an eigendecomposition as follows:

\begin{equation*}
    L_{norm} = U \Lambda U^{-1}
\end{equation*}

where the columns of \(U=[u_1, u_2, ... , u_{\vert\mathcal{V}\vert}]\) are the eigenvectors of \(L_{norm}\). It follows that, 

\begin{equation}\label{Asym}
    A_{norm} = I - U \Lambda U^{-1} = U (I - \Lambda) U^{-1} = \sum_{i=1}^{\vert\mathcal{V}\vert}u_iu_i^T(1-\lambda_i)
\end{equation}

Thus, in general we have:
\begin{equation}\label{Asymk}
    \sum_{k=0}^{m}\theta_kA^k_{norm} = \sum_{i=1}^{\vert\mathcal{V}\vert}u_iu_i^T\sum_{k=0}^{m}\theta_k(1-\lambda_i)^k
\end{equation}

The continuous functions \(\phi_1(x) = x\) and \(\phi_2(\lambda) = \sum_{k=0}^{m}\theta_k(1-\lambda)^k\) in \ref{alpha} can hence be approximated to an arbitrary accuracy by sufficiently wide MLPs by virtue of standard universal approximation results \citep{HORNIK1989359}.

(2) We need to prove that there exist functions \(\phi_1\) and \(\phi_2\) such that  \(\phi_1\big(\sum_{k=1}^{n}\sigma^g_k[i,j]\cdot\phi_2(\lambda^g_k)\big) \approx f(SPD^g[i,j])\) for every graph \(g\) with adjacency matrix \(A \in \mathbb{G}_n\) and every node-pair \((i,j)\) in \(g\).

If we can ensure that we can choose \(\phi_2\) such that it is possible to map every value \(\sum_{k=1}^{n}\sigma^g_k[i,j]\cdot\phi_2(\lambda^g_k)\) to the unique value \(f(SPD^g[i,j])\), then we can simply define \(\phi_1: \mathbb{R} \rightarrow \mathbb{R}\) as the piecewise-linear continuous function:

\begin{equation}\label{phi1}
    \phi_1\Big(\sum_{k=1}^{n}\sigma^g_k[i,j]\cdot\phi_2(\lambda^g_k)\Big) = \phi_1\Big(\sum_{k=1}^{n}u^g_k[i]u^g_k[j]\cdot\phi_2(\lambda^g_k)\Big)= f(SPD^g[i,j])
\end{equation}

which can be approximated by a sufficiently wide MLP. The proof then amounts to showing the existence of such a \(\phi_2\).

\textbf{Notation:} With a slight abuse of notation, let \(v^g_{ij} = [u^g_1[i]u^g_1[j], u^g_2[i]u^g_2[j], ... , u^g_n[i]u^g_n[j]]^T\). Further denote by \(\phi_2^g\) the vector \([\phi_2(\lambda^g_1), ... , \phi_2(\lambda^g_n)]^T\), so that \(\sum_{k=1}^{n}u^g_k[i]u^g_k[j]\cdot\phi_2(\lambda^g_k) = {v^g_{ij}}^T\phi^g_2\).

Now, let \(\phi_2(\lambda) = \sum_{k=0}^m \theta_k(1-\lambda)^k\) for some \(m \in \mathbb{Z}\).

\textbf{Case 1:} \(v^g_{ij} = v^g_{pq}\) for some graph \(g\) and node-pairs \((i,j)\) and \((p,q)\) in \(g\).
\newline
In this case, by virtue of \ref{Asymk}, \(SPD^g[i,j] = SPD^g[p,q]\) since \(A^k_{norm}[i,j]=A^k_{norm}[p,q]\), which implies that there is a \(k\)-hop path from \(i\) to \(j\) iff there is a \(k\)-hop path from \(p\) to \(q\). Moreover, in this case \({v^g_{ij}}^T\phi^g_2 = {v^g_{pq}}^T\phi^g_2\).

\textbf{Case 2:} \(v^g_{ij} \neq v^g_{pq}\) for some graph \(g\) and node-pairs \((i,j)\) and \((p,q)\) in \(g\).
\newline
If \(f(SPD^g[i,j]) = f(SPD^g[p,q])\), we simply set \(\phi_1({v^g_{ij}}^T\phi^g_2) = f(SPD^g[i,j])\) and \(\phi_1({v^g_{pq}}^T\phi^g_2) = f(SPD^g[p,q])\).

If \(f(SPD^g[i,j]) \neq f(SPD^g[p,q])\) and \({v^g_{ij}}^T\phi^g_2 \neq {v^g_{pq}}^T\phi^g_2\), we are done. If however, \({v^g_{ij}}^T\phi^g_2 = {v^g_{pq}}^T\phi^g_2\), choose an index \(r\) such that \(v^g_{ij}[r] \neq v^g_{pq}[r]\). We can then perturb \(\theta_r\) such that \({v^g_{ij}}^T\phi^g_2\) becomes unequal to \({v^g_{pq}}^T\phi^g_2\), in which case we again can set \(\phi_1({v^g_{ij}}^T\phi^g_2) = f(SPD^g[i,j])\) and \(\phi_1({v^g_{pq}}^T\phi^g_2) = f(SPD^g[p,q])\). We are assured that such a perturbation exists without violating the continuity constraints of \(\phi_1\) in \ref{phi1} for other graphs and node-pairs in \(\mathbb{G}_n\) as the set of such constraints is finite and bounded.

\textbf{Case 3:} \(v^{g_1}_{ij} = v^{g_2}_{pq}\) for some graphs \(g_1, g_2\) and node-pairs \((i,j)\) in \(g_1\) and \((p,q)\) in \(g_2\).
\newline
If \(f(SPD^{g_1}[i,j]) = f(SPD^{g_2}[p,q])\), we simply set \(\phi_1({v^{g_1}_{ij}}^T\phi^{g_1}_2) = f(SPD^{g_1}[i,j])\) and \(\phi_1({v^{g_2}_{pq}}^T\phi^{g_2}_2) = f(SPD^{g_2}[p,q])\).

If \(f(SPD^g[i,j)] \neq f(SPD^g[p,q])\) and \({v^{g_1}_{ij}}^T\phi^{g_1}_2 \neq {v^{g_2}_{pq}}^T\phi^{g_2}_2\), we are done. If however, \({v^{g_1}_{ij}}^T\phi^{g_1}_2 = {v^{g_2}_{pq}}^T\phi^{g_2}_2\), choose an index \(r\) such that \(\lambda^{g_1}_r \neq \lambda^{g_2}_r\) (which exists necessarily since otherwise \(SPD^{g_1}[i,j] = SPD^{g_2}[p,q]\) due to \ref{Asymk}). We can then perturb \(\theta_r\) such that \({v^{g_1}_{ij}}^T\phi^{g_1}_2\) becomes unequal to \({v^{g_2}_{pq}}^T\phi^{g_2}_2\), in which case we again can set \(\phi_1({v^{g_1}_{ij}}^T\phi^{g_1}_2) = f(SPD^{g_1}[i,j])\) and \(\phi_1({v^{g_2}_{pq}}^T\phi^{g_2}_2) = f(SPD^{g_2}[p,q])\). Similarly to Case 2 above, we are assured that such a perturbation exists without violating the continuity constraints of \(\phi_1\) in \ref{phi1} for other graphs and node-pairs in \(\mathbb{G}_n\).

\textbf{Case 4:} \(v^{g_1}_{ij} \neq v^{g_2}_{pq}\) for some graphs \(g_1, g_2\) and node-pairs \((i,j)\) in \(g_1\) and \((p,q)\) in \(g_2\).
\newline
If \(f(SPD^{g_1}[i,j]) = f(SPD^{g_2}[p,q])\), we simply set \(\phi_1({v^{g_1}_{ij}}^T\phi^{g_1}_2) = f(SPD^{g_1}[i,j])\) and \(\phi_1({v^{g_2}_{pq}}^T\phi^{g_2}_2) = f(SPD^{g_2}[p,q])\).

If \(f(SPD^g[i,j]) \neq f(SPD^g[p,q])\) and \({v^{g_1}_{ij}}^T\phi^{g_1}_2 \neq {v^{g_2}_{pq}}^T\phi^{g_2}_2\), we are done. If however, \({v^{g_1}_{ij}}^T\phi^{g_1}_2 = {v^{g_2}_{pq}}^T\phi^{g_2}_2\), choose an index \(r\) such that \(v^{g_1}_{ij}[r] \neq v^{g_2}_{pq}[r]\). We can then perturb \(\theta_r\) such that \({v^{g_1}_{ij}}^T\phi^{g_1}_2\) becomes unequal to \({v^{g_2}_{pq}}^T\phi^{g_2}_2\), in which case we again can set \(\phi_1({v^{g_1}_{ij}}^T\phi^{g_1}_2) = f(SPD^{g_1}[i,j])\) and \(\phi_1({v^{g_2}_{pq}}^T\phi^{g_2}_2) = f(SPD^{g_2}[p,q])\). Similarly to Case 2 and 3 above, we are assured that such a perturbation exists without violating the continuity constraints of \(\phi_1\) in \ref{phi1} for other graphs and node-pairs in \(\mathbb{G}_n\).

Thus, we have proven that the continuous function \(\phi_2(\lambda) = \sum_{k=0}^m \theta_k(1-\lambda)^k\) for some \(m \in \mathbb{Z}\) can be chosen to ensure the continuity of \(\phi_1\) defined in \ref{phi1}. Both functions can hence be approximated by MLPs to an arbitrary accuracy due to \citet{HORNIK1989359}.

\section{Experiment Details}\label{expdetails}
\subsection{Description of Datasets}
Details specific to each dataset used in our results are presented in Table \ref{datasets-table}.

\begin{table}[h]
  \caption{Overview of datasets used in our work}
  \label{datasets-table}
  \centering
  \begin{tabular}{lcccccc}
    \toprule
    Dataset & \#graphs & \#nodes & \#edges & Directed & Level & Task\\
    \midrule
    ZINC & 12,000 & \(\sim\)23.2 & \(\sim\)24.9 & No & Graph & Regression\\
    MNIST & 70,000 & \(\sim\)70.6 & \(\sim\)564.5 & Yes & Graph & 10-Class Classification\\
    CIFAR10 & 60,000 & \(\sim\)117.6 & \(\sim\)941.2 & Yes & Graph & 10-Class Classification\\
    PATTERN & 14,000 & \(\sim\)118.9 & \(\sim\)3039.3 & No & Node& Binary Classification\\
    CLUSTER & 12,000 & \(\sim\)117.2 & \(\sim\)2150.9 & No & Node & 6-Class Classification\\
    \midrule
    Peptides-func & 15,535 & \(\sim\)150.9 & \(\sim\)307.3 & No & Graph & 10-Class Multilabel\\
     &  &  &  &  &  & Classification\\
    Peptides-struct & 15,535 & \(\sim\)150.9 & \(\sim\)307.3 & No & Graph & 11-Task Regression\\
    \bottomrule
  \end{tabular}
\end{table}

\subsection{Hyperparameters}
Due to a limited training budget we could not perform a large scale hyperparameter search and mostly borrowed configurations from \citet{GRIT} and \citet{GraphGPS}, keeping in line with the parameter budgets, optimizer (AdamW with \(betas=(0.9, 0.99)\) and \(eps=1e-8\)) and learning rate schedule (warmup+cosine annealing) used therein. Notably, due to the simple attention mechanism of \textsc{Eigenformer} (without feature attention), we could use deeper/wider models for training. Further details are included in Table \ref{gnnbhp-table} and \ref{lrgbhp-table} (SSR = Signed-square-root. For the notation A/B, A=metric without feature attention and B=metric with feature attention).

\begin{table}
  \caption{Best hyperparameter configurations for the benchmarks from \citet{Benchmarking}}
  \label{gnnbhp-table}
  \centering
  \begin{tabular}{lccccc}
    \toprule
    Hyperparameter & ZINC & MNIST & CIFAR10 & PATTERN & CLUSTER\\
    \midrule
    \# Transformer Layers & 12 & 3 & 3 & 10 & 16 \\ 
    \# Heads & 8 & 8 & 8 & 8 & 8 \\
    \(\psi\) & SSR & SSR & Identity & SSR & SSR \\
    Node feature embedding & 128 & - & - & - & - \\
    Edge feature embedding & 128 & - & - & - & - \\
    Hidden dim. & 72/56 & 64/56 & 64/56 & 64/64 & 56/56 \\
    \(\phi_1\) Hidden dim. & 28/28 & 28/24 & 28/24 & 36/28 & 28/24 \\
    \(\phi_2\) Hidden dim. & 28/28 & 28/24 & 28/24 & 36/28 & 28/24 \\
    Dropout & 0 & 0 & 0 & 0 & 0\\
    Attention dropout & 0.2 & 0.2 & 0.2 & 0.5 & 0.5 \\
    Graph pooling & sum & mean & mean & - & - \\
    \midrule
    Batch size & 128/512 & 512 & 512 & 16/24 & 16 \\
    Learning rate & 0.001 & 0.001 & 0.001 & 0.0005 & 0.0005 \\
    \# Epochs & 2000 & 200 & 200 & 100 & 100 \\
    \# Warmup epochs & 50 & 5 & 5 & 5 & 5 \\
    Weight decay & 1e-5 & 1e-5 & 1e-5 & 1e-5 & 1e-5 \\
    \midrule
    \# Parameters (w/o feat) & 509849 & 106746 & 106874 & 476929 & 486006 \\
    \# Parameters (w/ feat) & 479481 & 110234 & 110346 & 472321 & 479734 \\
    \midrule
    \# Epoch time (s) & 21/23 & 198/222 & 793/840 & 571/619 & 550/582 \\
    \# Accelerator (NVIDIA) & A10G & A10G & A10G & A10G & A10G \\
    \# Peak memory (GB) & 19.44/20.54 & 19.95/19.19 & 20.39/19.06 & 16.38/21.17 & 17.9/17.53 \\
    \bottomrule
  \end{tabular}
\end{table}

\begin{table}
  \caption{Best hyperparameter configurations for the benchmarks from \citet{LRGB}}
  \label{lrgbhp-table}
  \centering
  \begin{tabular}{lcc}
    \toprule
    Hyperparameter & Peptides-func & Peptides-struct\\
    \midrule
    \# Transformer Layers & 10 & 10\\ 
    \# Heads & 8 & 8\\
    \(\psi\) & SSR & SSR \\
    Hidden dim & 80/64 & 80/64\\
    \(\phi_1\) Hidden dim & 32/28 & 32/28 \\
    \(\phi_2\) Hidden dim & 32/28 & 32/28 \\
    Dropout & 0 & 0\\
    Attention dropout & 0.5 & 0.5\\
    Graph pooling & mean & mean\\
    \midrule
    Batch size & 12/16 & 14/14\\
    Learning rate & 0.0003 & 0.0003\\
    \# Epochs & 200 & 200\\
    \# Warmup epochs & 5 & 5\\
    Weight decay & 1e-5 & 1e-5\\
    \midrule
    \# Parameters (w/o feat) & 510570 & 510651\\
    \# Parameters (w/ feat) & 510651 & 478779 \\
    \midrule
    \# Epoch time (s) & 1740/2736 & 2541/2700 \\
    \# Accelerator (NVIDIA) & H100 & H100 \\
    \# Peak memory (GB) & 60.76/75.76 & 70.77/66.69 \\
    \bottomrule
  \end{tabular}
\end{table}


\end{document}